\documentclass[letterpaper, 10 pt, conference]{ieeeconf}  

\IEEEoverridecommandlockouts                              

\usepackage{wrapfig}
\usepackage{subcaption}
\usepackage{amsmath}
\usepackage{amssymb}
\usepackage{float}
\usepackage{cleveref}
\usepackage{caption}
\usepackage{makecell}
\usepackage{diagbox}
\usepackage{tabularx}
\usepackage[table]{xcolor}
\usepackage{multirow}
\usepackage{cite}

\usepackage{graphicx}
\graphicspath{{./img}}

\title{\LARGE \bf
Multi-Robot Collaboration through Reinforcement Learning and Abstract Simulation
}

\author{Adam Labiosa$^{1}$ and Josiah P. Hanna$^{1}$
\thanks{$^{1}$University of Wisconsin-Madison,
        Madison, WI, USA. Correspondence to
        {\tt\small labiosa@wisc.edu}}%
}

\begin{document}

\maketitle
\thispagestyle{empty}
\pagestyle{empty}

\begin{abstract}
Teams of people coordinate to perform complex tasks by forming abstract mental models of world and agent dynamics. The use of abstract models contrasts with much recent work in robot learning that uses a high-fidelity simulator and reinforcement learning (RL) to obtain policies for physical robots. Motivated by this difference, we investigate the extent to which so-called \textit{abstract simulators} can be used for multi-agent reinforcement learning (MARL) and the resulting policies successfully deployed on teams of physical robots. An abstract simulator models the robot's target task at a high-level of abstraction and discards many details of the world that could impact optimal decision-making. Policies are trained in an abstract simulator then transferred to the physical robot by making use of separately-obtained low-level perception and motion control modules. We identify three key categories of modifications to the abstract simulator that enable policy transfer to physical robots: simulation fidelity enhancements, training optimizations and simulation stochasticity. We then run an empirical study with extensive ablations to determine the value of each modification category for enabling policy transfer in cooperative robot soccer tasks. We also compare the performance of policies produced by our method with a well-tuned non-learning-based behavior architecture from the annual RoboCup competition and find that our approach leads to a similar level of performance. Broadly we show that MARL can be use to train cooperative physical robot behaviors using highly abstract models of the world. 

\label{sec:abstract}
\end{abstract}


\section{Introduction}
\label{sec:introduction}

Teamwork enables groups of agents (humans, animals, and robots) to accomplish complex tasks more efficiently than any individual can alone. For example, orca whales use teamwork to generate waves and push prey off of ice \cite{pitman2012cooperative}.
Likewise, autonomous robots can accomplish far more when they work as a team.
For example, drones patrolling for wildfires can cover far more area if they coordinate the area they cover \cite{momeni2023multi}.
Multi-agent reinforcement learning (MARL) is a promising approach for obtaining such cooperative behaviors for autonomous robots with a success-defining reward function and task interaction time.

While MARL is a general approach to developing cooperative behaviors, current MARL algorithms are data inefficient and many success stories \cite{silver2016mastering, silver2017mastering, schrittwieser2020mastering, vinyals2019grandmaster, berner2019dota, baker2019emergent} have taken place exclusively in simulated environments where data is comparatively cheap to collect.
Unfortunately, this scale of data collection is currently infeasible on physical robots.

\begin{figure}[t]
\vspace{3mm}
\centering
\begin{subfigure}[t]{0.39\columnwidth}
    \centering
    \includegraphics[width=0.9\textwidth]{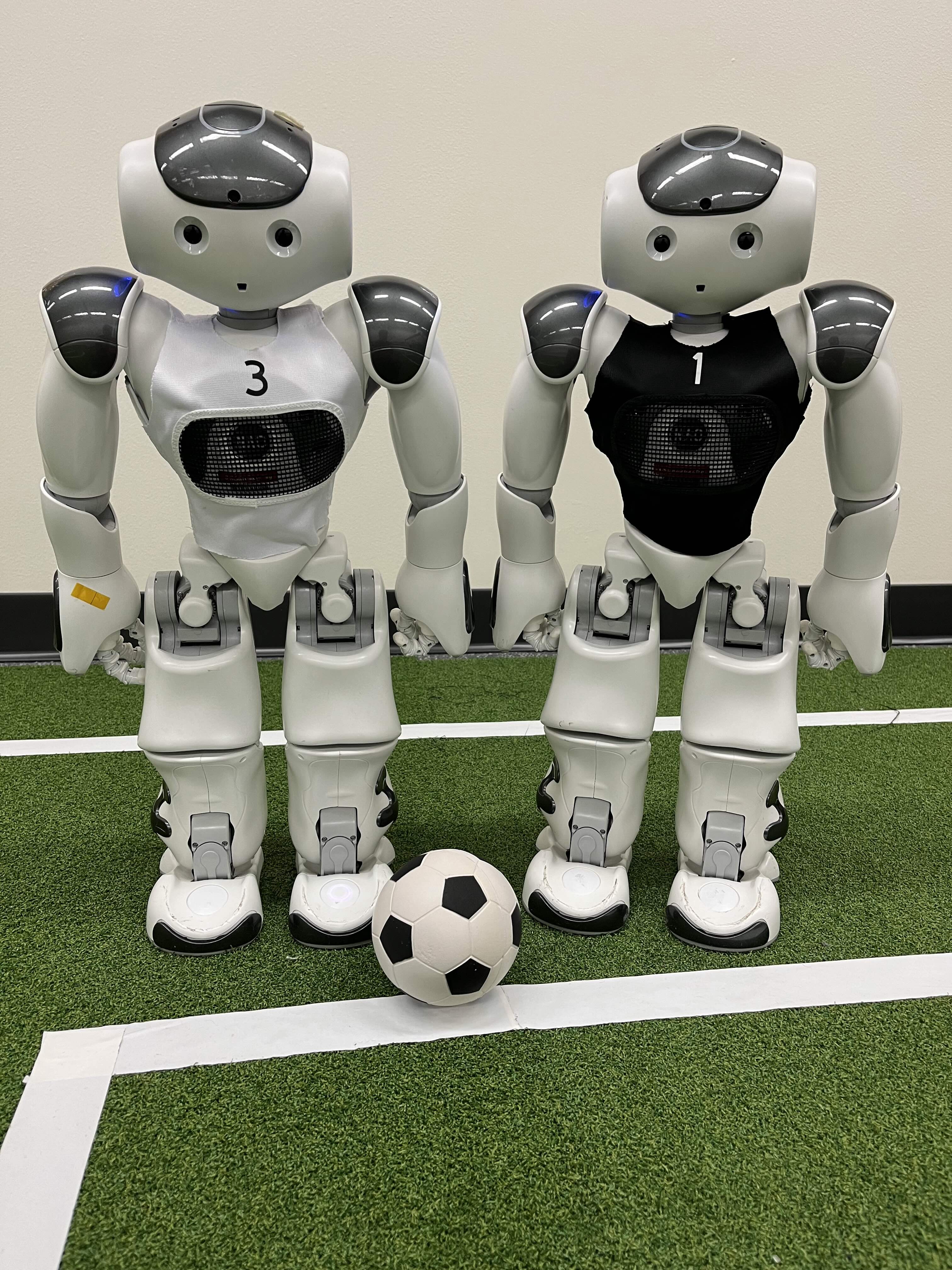}
    \caption{Two NAO humanoid robots used in physical experiments.}
    \label{fig:NAOs}
\end{subfigure}
\hfill
\begin{subfigure}[t]{0.59\columnwidth}
    \centering
    \includegraphics[width=0.9\textwidth]{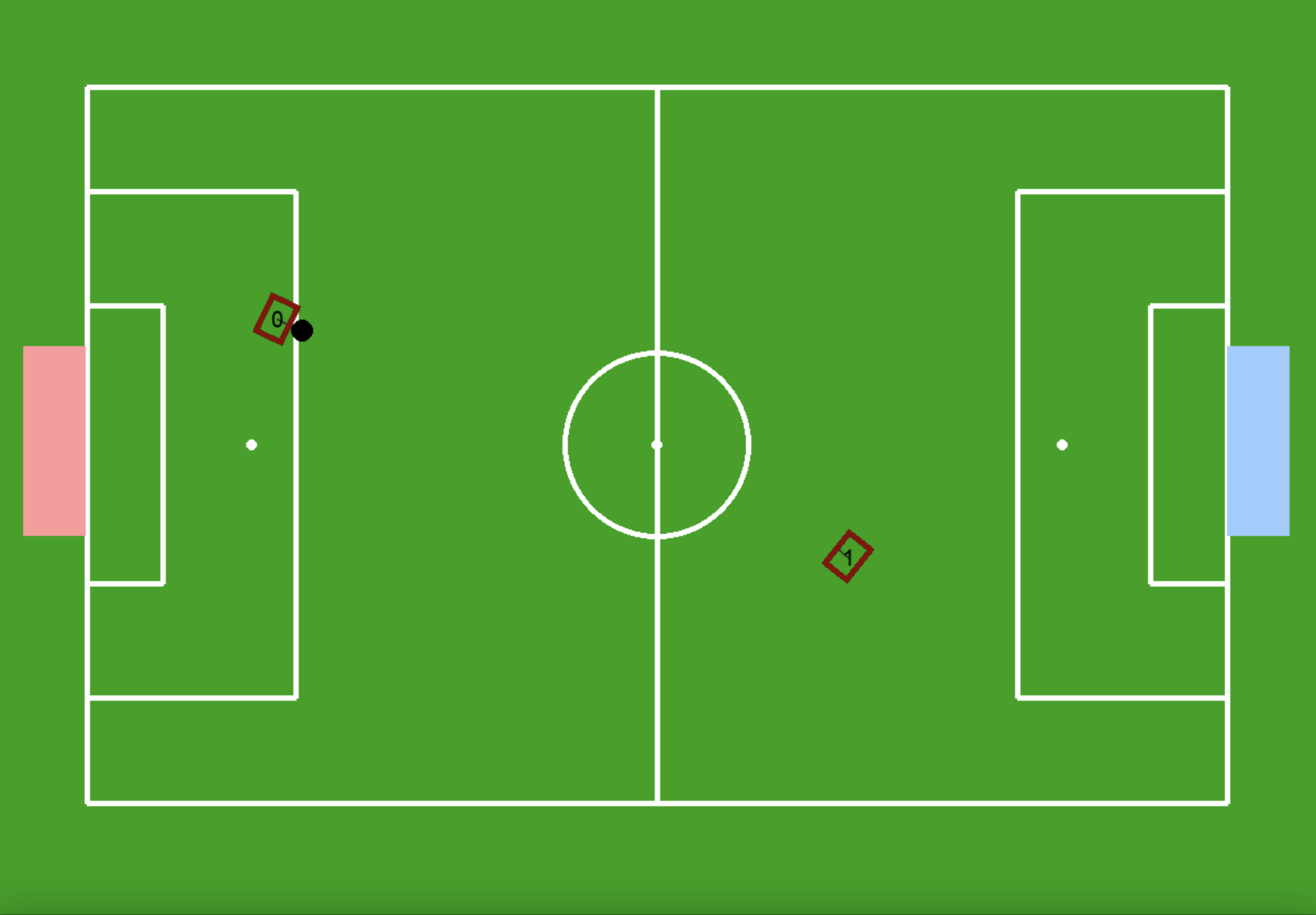}
    \caption{Visualization of our abstract simulation used to train policies before deployment on robots. Robots are the small red boxes and the ball is the black circle.}
    \label{fig:abSimSoccer}
\end{subfigure}
\caption{Comparison of the physical NAOv6 used in experiments to the abstract simulation we use to train the policies to control high-level cooperative behaviors.}
\vspace{-2mm}
\end{figure}

To take advantage of simulation for physical robots, much recent work has gone toward developing realistic high-fidelity robotics simulators \cite{makoviychuk2021isaac, todorov2012mujoco, serban2018chrono} to reduce the inevitable \textit{reality gap} between simulation and reality.
Complementing this research, in this work we study if high-fidelity simulation is necessary for developing cooperative control policies for physical robots.
We draw inspiration from human planning \cite{ho2022people} and train in an \textit{abstract} simulation which models the world at a low-fidelity, coarse level of detail (\Cref{fig:abSimSoccer}).

In addition to being a step toward imbuing robots with abstract reasoning capabilities, abstract simulators offer several benefits for learning cooperative robotic behaviors. First, in many domains, they will be easier to create and require much less domain knowledge. Second, abstract simulations simplify the world model and thus reduce the need for complex and slow physics calculations. Finally, cooperation requires mostly high-level reasoning, so extensive low-level modeling of the target domain may be unnecessary.
With this motivation in mind, our work seeks to answer the question:
\begin{center}
\textit{Can a team of robots learn cooperative behaviors using MARL within an abstract simulation of their task such that these behaviors transfer to physical robots?}
\end{center}
In this paper, we answer this question affirmatively. In doing so, this work makes the following contributions:
\begin{enumerate}
\item We demonstrate that multi-agent policies trained in an abstract simulator can successfully transfer to legged physical robots.
\item We identify and categorize key considerations for improving simulator realism and MARL training that enable zero-shot policy transfer to physical robots. A particularly surprising finding was that \textit{decreasing} simulator fidelity sometimes \textit{increased} sim2real transfer by promoting more effective MARL in simulation.
\item We conduct an extensive ablation study to determine the critical attributes that facilitate the use of abstract simulation to train multi-robot control policies.
\item We demonstrate that our approach produces policies for teams of two robots that perform on par with extensively tuned behaviors developed by recent champions of the RoboCup Standard Platform League competition.\footnotemark[2]
\end{enumerate}
With these contributions, we provide a framework for developing transferable multi-agent policies using abstract simulation, potentially reducing the need for expensive and time-consuming real-world training in multi-robot applications. 

\footnotetext[2]{We note that the RoboCup SPL competition presents a greater challenge than the specific evaluation scenarios we designed to test multi-agent cooperation and thus, we do not claim that the trained MARL policies alone would match the performance of an SPL team.}


\section{Related Work}

In this section, we discuss prior work on abstraction in robotics and sim2real.

\label{sec:relatedWork}

\subsection{Abstraction for Robotic Reinforcement Learning}

Abstraction is a common technique used to simplify the learning process for RL agents \cite{li2006towards}. Prior to deep RL, abstraction was used for navigation tasks for various wheeled and snake-like robots \cite{provost2004self, shibata2006learning, ito2007autonomous, frommberger2010structural}. Other works extend abstraction for robotics into deep RL \cite{ito2016reinforcement, veerapaneni2020entity, agarwal2023legged, geles2024demonstrating} but do not use abstract simulation to the extent we do in this work nor do they consider multi-agent domains. Truong et al.\ \cite{truong2023rethinking} uses an abstract simulation to accomplish navigation tasks but also does not consider the multi-agent case. Zhang et al.\ \cite{zhang2024back} uses abstract simulation for multi-agent drones but does not consider the more complex domain of legged robots.

Another approach to abstraction is to use a hierarchy where low and high-level control are trained separately. Stulp et al.\ \cite{stulp2011hierarchical} uses pretrained low-level motion primitives like we do in our work, but they do not use deep RL and it takes place in a simpler single-agent grasping domain. Ma et al.\ \cite{ma2020hierarchical} also uses motion primitives for hierarchical RL and considers a multi-agent domain but does not use physical robots. Nachum et al.\ \cite{nachum2019multi} uses hierarchy to train motion control and then high-level control in a multi-agent domain. Their work considers a simpler domain for pushing boxes with quadrupeds and uses high-fidelity simulation for both low- and high-level policy training.

Some works use low and high-fidelity simulation to balance fast training and real world performance \cite{cutler2015real, qiu2021low, beard2022black, agrawal2023reinforcement, bhola2023multi, leguizamo2022deep, truong2021learning}. Our work only uses low-fidelity training and explores the multi-agent space while these works are focused on single-agent training.

\subsection{Sim2real Reinforcement Learning}

Reinforcement learning methods generally require large amounts of data to exhibit desirable behaviors. For example, learning to play Dota 2 required 10 months of training on thousands of CPUs and hundreds of GPUs \cite{berner2019dota}. This computational scale is infeasible for physical robots due to the prohibitive monetary and time costs. Therefore, many works adopt the sim2real paradigm -- training policies in simulation and transferring the fixed policies to physical robots \cite{zhao2020sim, salvato2021crossing}.

Sim2real transfer has produced intelligent robot behavior across a wide variety of domains \cite{kaufmann2023champion, hwangbo2017control, cao2019target, carlucho2018adaptive, hester2013texplore, mahmood2018benchmarking, falco2018policy, hazara2019transferring, chalvatzaki2019learn, loquercio2021learning}. In the domain of robot soccer, high-fidelity sim2real has achieved vision-to-motor control, complex multi-agent behaviors and multi-robot teamwork \cite{haarnoja2024learning, tirumala2024learning, duan2012multi, park2001modular}. Despite the successes of high-fidelity sim2real, the challenge of the reality gap remains. Popular methods to bridge this gap include domain randomization \cite{tobin2017domain}, and precise simulation-to-reality matching \cite{kaspar2020sim2real, hanna2021grounded, tan2018sim}. 

Many works utilize sim2real transfer to enable multi-agent behaviors on real robots \cite{blumenkamp2022framework, de2020deep, cui2024distributed}. Some works \cite{candela2022transferring, zhao2020towards} employ sim2real transfer to deploy multi-agent policies on real robots, but they focus on end-to-end deployment which generally requires a high-fidelity simulation. 

Sim2real RL can be considered as an instance of model-based RL. Model-based single-agent RL has been applied to learn on physical robots by learning a world model and training agents in that model \cite{djeumou2023learn, huang2022trade, wu2023daydreamer, thuruthel2018model, zhang2019solar, hester2012rtmba, becker2020learning, martinez2015safe, chalvatzaki2019learn, lambert2019low, imanberdiyev2016autonomous}.  Model-based RL has also been applied to multi-agent tasks for increased multi-agent performance \cite{van2021model, zhang2020model, pasztor2021efficient, sessa2022efficient, egorov2022scalable}. These works focused on learning world models as opposed to using a crafted simulation and do not evaluate on physical robots.


\section{Background}
\label{sec:background}

We formalize the MARL problem as solving a stochastic game, also known as a Markov game. A stochastic game is defined by the tuple $(\mathcal{S}, \mathcal{A}_1, \ldots, \mathcal{A}_n, p, r_1, \ldots, r_n)$, where $n$ is the number of agents; $\mathcal{S}$ is the set of all possible states of the environment; and $\mathcal{A}_i$ is the set of actions available to agent $i$. The joint action space $\mathcal{A}$ is defined as $\mathcal{A} = \mathcal{A}_1 \times \ldots \times \mathcal{A}_n$, representing all possible combinations of actions taken by the agents. $p(s'|s, a_1, \ldots, a_n)$ is the probability of transitioning to state $s'$ given the current state $s$ and the joint action $(a_1, \ldots, a_n)$ taken by all agents. $r_i: \mathcal{S} \times \mathcal{A} \rightarrow \mathbb{R}$ is the reward function for agent $i$, which depends on the current state and the joint action. In a stochastic game, the goal of each agent $i$ is to learn a policy $\pi_i: \mathcal{S} \rightarrow \mathcal{A}_i$ that maximizes its expected cumulative reward. This work focuses on a cooperative multi-agent setting. In cooperative games all agents share the same reward function and aim to maximize this shared cumulative reward.


\section{Robot Hardware and Simulation}

Before presenting our methodology for crossing the abstract simulator reality gap, we describe the robot hardware and simulation set-up that we will use.
We present these platforms to ground the presentation of our methodology in \Cref{sec:methodology}.

\subsection{Robot Hardware and Behavior Architecture}
\label{sec:robothardware}

\begin{figure}
\vspace{3mm}
    \centering
    \includegraphics[width=0.75\linewidth]{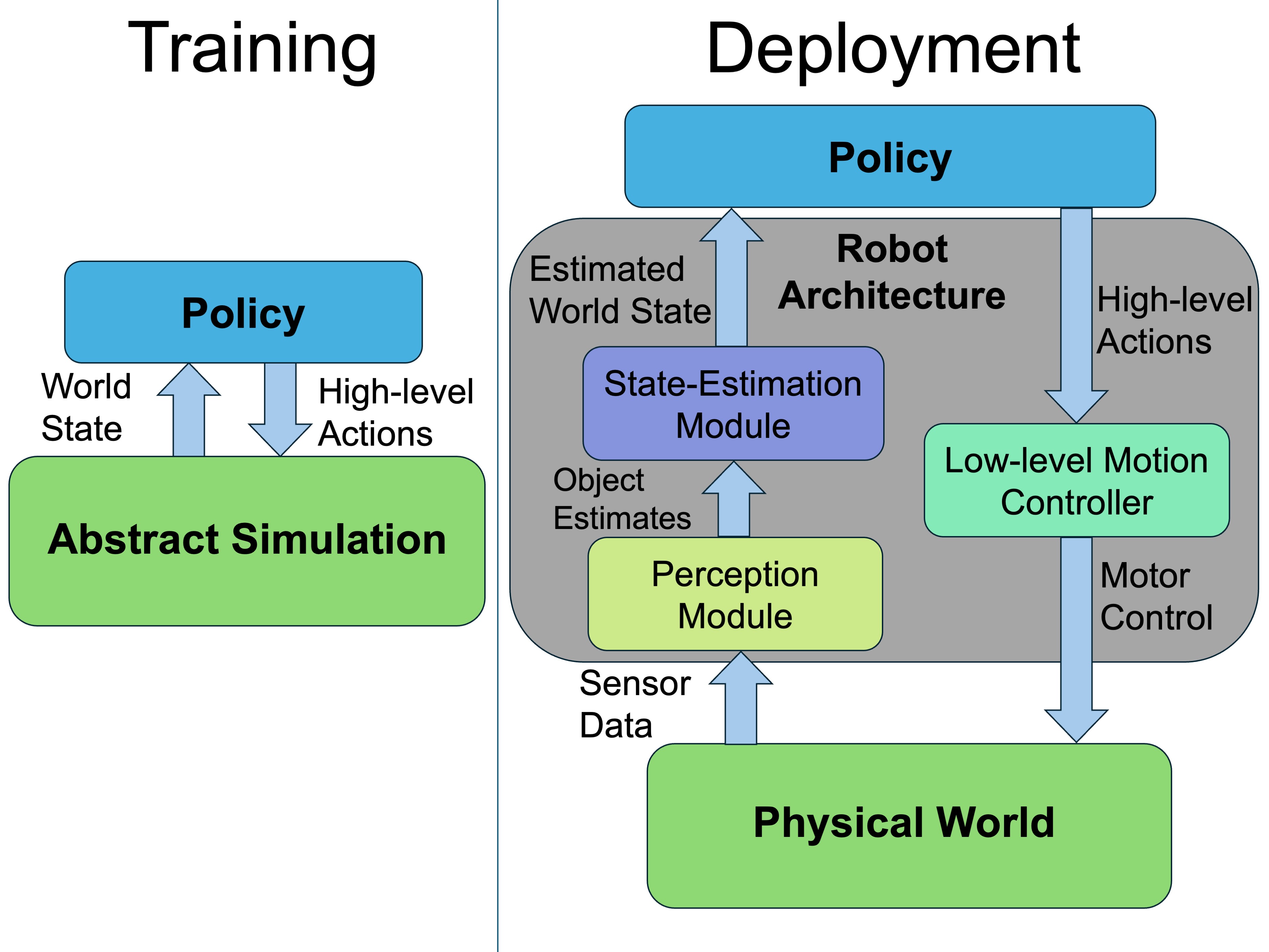}
    \caption{Visualization of our training and deployment interaction models. Exact world state is given during training and world state is estimated by the robot architecture modules during deployment. Similarly, in simulation the policy exactly controls high-level actions while during deployment they are passed to a motion controller which controls motor outputs.}
    \label{fig:trainingDeploymentStack}
    \vspace{-2mm}
\end{figure}

In this work, we use the NAOv6 humanoid robot (\Cref{fig:NAOs}). 
The NAO has 25 degrees of freedom and 2 cameras, foot pressure sensors, and IMUs for perception.
All computation runs onboard the robot using a 4-core Intel processor.

Our work focuses on enabling MARL to learn high-level decision-making for the NAO in cooperative tasks.
To facilitate the use of high-level policies, we make use of separately implemented low-level perception and motion capabilities. We train without an end-to-end approach and build upon the architecture developed by the RoboCup team, B-Human \cite{bhuman_code_release}.
This behavior architecture provides the full capabilities required to participate in the RoboCup Standard Platform League competition including modules that handle real-time object detection, localization, and motor control and high-level modules for robot behaviors (\Cref{fig:trainingDeploymentStack}).
Our focus in this work is to learn a policy that replaces the high-level behavior modules.

The state and action spaces are designed to maximize the benefits of the abstract simulation and the existing robot architecture. The state space $\mathcal{S}$ includes egocentric representations of objects around the field including the ball, teammate and opponent. The agents action space $\mathcal{A}$ is given as longitudinal movement, lateral movement, angular rotation, and a threshold output for kick and stand.

\subsection{Abstract Simulation}

Since we focus on investigating the use of high-level abstract simulators for MARL, in this section, we introduce the abstract simulator (AbstractSim -- \Cref{fig:abSimSoccer}) that we used in this study to ground the following sections. We note that AbstractSim is not part of our contribution and instead it was developed independently for single-agent RL use in the RoboCup competition \cite{labiosa2024reinforcement}. We use it here as a case study to evaluate the role of abstract simulations in complex, multi-agent tasks.
AbstractSim models the state and transition dynamics of the world without explicitly modeling how the physical robot estimates the state or carries out actions. 
Object representations are also simplified with the bipedal robots represented with rectangular collision volumes and no agent-to-agent collision dynamics.
Both of these design decisions impact the dynamics of both the robot and soccer ball we use in our experiments. At every timestep the robot moves precisely in the direction indicated by the policy, abstracting away complexities such as momentum and gait and ignoring all dynamics of legged movement. Ball dynamics are also simplified. The ball moves at a consistent speed and direction upon contact and decelerates at a fixed rate.

\begin{figure}
\vspace{3mm}
    \centering
    \includegraphics[width=0.7\linewidth]{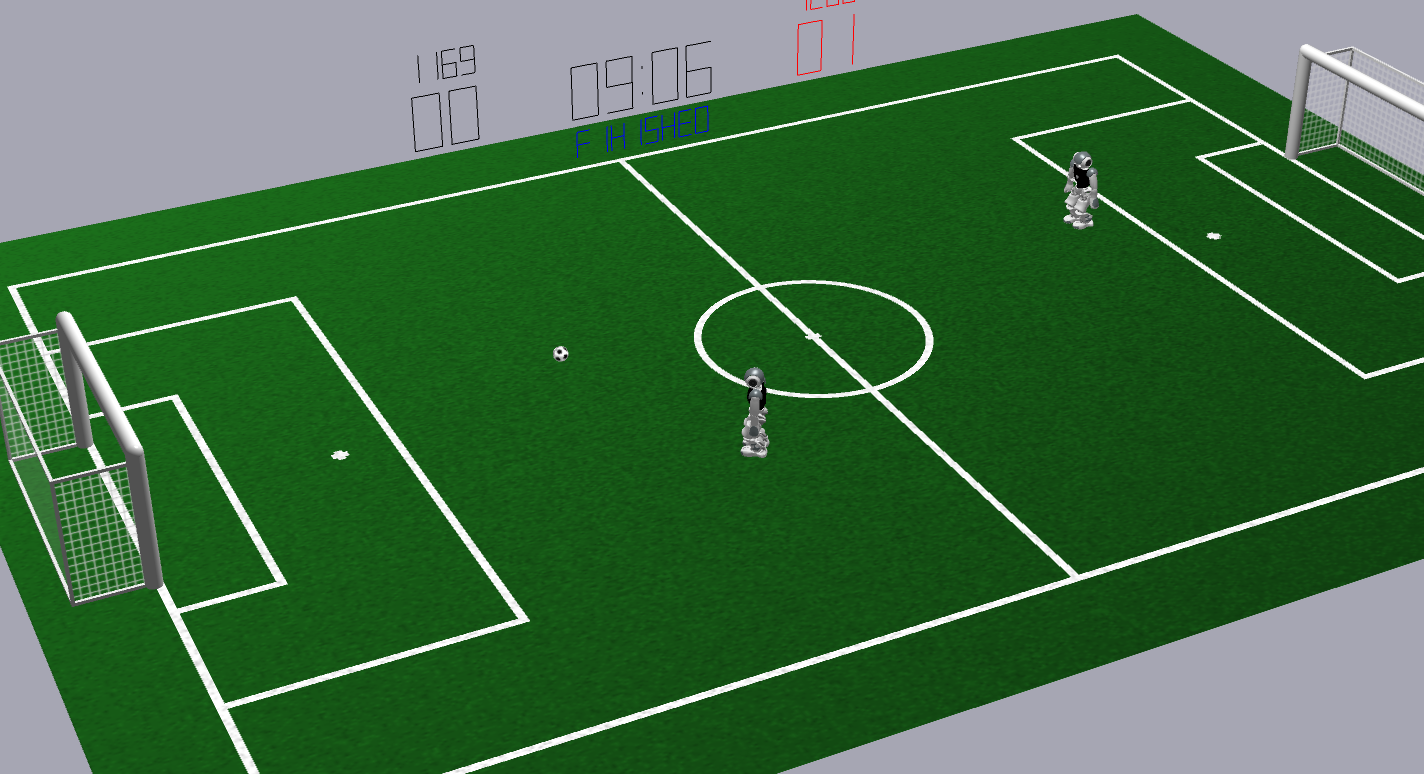}
    \caption{High-fidelity simulation developed by the B-Human RoboCup Team. Used for simulation experiments. Physics are based on the Open Dynamics Engine \cite{smith2005open}.}
    \label{fig:SimRobot}
    \vspace{-3mm}
\end{figure}

\begin{figure*}[ht]
\vspace{3mm}
    \centering
    
    \begin{subfigure}[t]{0.48\linewidth}
        \scriptsize
        \centering
        \arrayrulecolor{gray!80}
        \rowcolors{2}{white}{gray!25}
        \begin{tabularx}{\linewidth}{|p{0.5cm}|XXXp{1.6cm}|}
        \hline
        & Full MARL (F) & Large Displacement (E) & Realistic Agent Size (T) & No Ball Noise (N) \\ \hline
        BS 1 & $\mathbf{0.90 \pm 0.2}$ & $0.70 \pm 0.3$ & $0.20 \pm 0.3$ & $0.10 \pm 0.2$ \\
        BS 2 & $\mathbf{1.00 \pm 0.0}$ & $0.70 \pm 0.4$ & $0.50 \pm 0.4$ & $0.90 \pm 0.2$ \\
        BS 3 & $\mathbf{0.80 \pm 0.3}$ & $0.60 \pm 0.4$ & $0.30 \pm 0.4$ & $0.30 \pm 0.4$ \\
        D 1 & $\mathbf{0.50 \pm 0.4}$ & $0.00 \pm 0.0$ & $0.40 \pm 0.4$ & $0.10 \pm 0.2$ \\
        D 2 & $\mathbf{1.00 \pm 0.0}$ & $0.80 \pm 0.3$ & $\mathbf{1.00 \pm 0.0}$ & $0.30 \pm 0.4$ \\
        D 3 & $0.80 \pm 0.3$ & $\mathbf{0.90 \pm 0.2}$ & $0.70 \pm 0.4$ & $0.00 \pm 0.0$ \\
        \hline
        \end{tabularx}
        \caption{Success Rates for Basic and Defender Physical Robot Experiments. Higher is better.}
    \end{subfigure}
    \hfill
    \begin{subfigure}[t]{0.48\linewidth}
        \centering
        \scriptsize
        \arrayrulecolor{gray!80}
        \rowcolors{2}{white}{gray!25}
        \begin{tabularx}{\linewidth}{|p{0.5cm}|XXXp{1.6cm}|}
        \hline
        & Full MARL (F) & Large Displacement (E) & Realistic Agent Size (T) & No Ball Noise (N) \\ \hline
        BS 1 & $45.78 \pm 8.1$ & $43.14 \pm 4.9$ & $\mathbf{39.50 \pm 1.0}$ & $60.00 \pm 0.0$ \\
        BS 2 & $\mathbf{24.30 \pm 8.8}$ & $24.86 \pm 6.4$ & $37.60 \pm 8.9$ & $39.44 \pm 5.8$ \\
        BS 3 & $43.75 \pm 5.6$ & $\mathbf{37.00 \pm 5.4}$ & $55.00 \pm 3.0$ & $49.67 \pm 9.2$ \\
        D 1 & $\mathbf{40.60 \pm 12.3}$ & $N/A$ & $54.25 \pm 4.8$ & $56.00 \pm 0.0$ \\
        D 2 & $23.50 \pm 3.0$ & $\mathbf{22.00 \pm 2.7}$ & $35.50 \pm 5.2$ & $45.67 \pm 20.3$ \\
        D 3 & $39.00 \pm 8.5$ & $\mathbf{32.22 \pm 3.5}$ & $\mathbf{32.22 \pm 3.5}$ & $N/A$ \\
        \hline
        \end{tabularx}
        \caption{Time to Score for Basic and Defender Physical Robot Experiments. Lower is better. Note that this is an average of successful runs.}
    \end{subfigure}
    
    \caption{Physical robot experiment results. F are full methods. E are fidelity enhancements. T are training optimizations. N are noise. Our full method is Full MARL. Each column provides an experiment location for either the Basic Soccer (BS) task or the Defender (D) task. Number of trials is 10. 95\% confidence intervals computed using the Student t-distribution.}
    \label{fig:physical_robot_results}

    \vspace{-1mm}
\end{figure*}

For MARL, AbstractSim offers two key benefits. First, the simplifications facilitate faster training which makes it easier to iterate on the MARL training approach (e.g., tuning the reward function and training set-up). In comparison to our high-fidelity simulation (\Cref{fig:SimRobot}) AbstractSim runs 30x as fast. Second, abstract robot models and environment dynamics are inherently simple and therefore easy to design. The key challenge with the significance difference between the simple AbstractSim dynamics and the complex real-world dynamics -- that we address next -- is that AbstractSim has a potentially large reality gap which hinders direct policy transfer. Concretely, with our initial version of the simulator, agents failed to learn effective cooperative behaviors or even simple single-agent behaviors. In the next section, we describe the key techniques that enable the use of MARL and our AbstractSim to train cooperative policies for the physical robots.

\section{Ingredients for Multi-agent Abstract Sim2Real}
\label{sec:methodology}

To successfully apply MARL in an abstract simulator, we identify three categories of changes to the simulator and training procedure that enable successful policy transfer.

\subsection{Simulation Fidelity Enhancements}

Some degree of realism is required to enable policy transfer even when the simulation is an intentionally abstract model of the world so we first consider modifications that improve the realism of world dynamics.
We found that increasing the realism of the simulation is beneficial only to the extent that the complexities do not hinder the learning of cooperative behaviors. Overly complex simulations can increase learning difficulty without improving policy transfer.
Therefore, we identify a small set of modifications that make key details more realistic. 

Concretely, we modify the simulated agents' speed to match the physical robots' speed. From the initial AbstractSim, we lowered the robot's angular and translational velocity movements to reduce the reality gap of simulator movement dynamics.

Contrary to the assumption that increased realism always enhances real-world performance, we observed that in some cases, decreasing certain elements of realism in the simulation actually improved the physical robots' performance. We term these counterintuitive simulation adjustments \textit{Training Optimizations}.

\subsection{Training Optimizations}

The objective of our method is to maximize performance on real robots, which requires both effective MARL in simulation and effective sim2real transfer. 
A counterintuitive observation is that some changes -- designed to increase realism and promote transfer -- led to MARL failing to learn teamwork in AbstractSim. Consequently, the policies we obtained were not performant on the physical robot.
We observe that too much realism can make the simulation unnecessarily complex and challenging, hindering the learning process. Intuitively, the agents do learn to maximize the single-agent behavior but due to complex dynamics, do not experience multi-agent trajectories and therefore do not learn multi-agent behavior. Therefore, it is sometimes beneficial to make changes that \textit{decrease} realism so as to balance cooperative trainability with transferability.

Concretely, we modified three aspects of our simulator to less closely match the physical world: robot size, goal size and the time to kick. First, we increased the robot size to about four times the size of a real NAO which increased the ball-robot contact distance. Second, we halved the goal size to force precision when kicking into the goal. Finally, we removed any time delay during kicking as it takes about one second for the physical robots to kick the ball. These modifications, while seemingly counterintuitive, seek to address the reality gap by encouraging the learning of more robust and cooperative policies that are less sensitive to the dynamics of the simulation environment.

\subsection{Simulation Stochasticity}

Even when it is beneficial to make an abstract simulator more realistic, there are limits to how realistic we can make the simulation without significantly increasing its complexity.
Accurately modeling complex real-world dynamics like robot-to-ball contact would require details models of the robots and world kinematics which would increase complexity and possibly hinder learning. Instead of increasing realism, we propose to increase robustness by adding noise into hard to model dynamics.
This type of randomization is sometimes used as part of domain randomization for sim2real.

Concretely, we considered two uses of noise with our platforms.
First, during training, we add noise to robot-ball contact during a push contact but not a kick, as our robots have precise kicking capabilities. We add uniform noise as we observe highly stochastic motion following contacts on the physical robots. 
We also explore the use of noise to encourage robustness to error in localization on the physical robot. We do so by adding noise to the observations of the robot during training.

By introducing noise into the simulation, we increase the variability of the environment and encourage the learning of more robust policies that are less sensitive to the uncertainties of the real-world.


\begin{figure*}[ht]
\centering
\vspace{3mm}
    \begin{subfigure}[b]{\linewidth}
        \scriptsize
        \centering
        \arrayrulecolor{gray!80} 
        \rowcolors{2}{white}{gray!25} 
        \begin{tabularx}{\linewidth}{|p{0.5cm}|XXXXXXXXXX|}
        \hline
        & Full MARL (F) & BHuman (F) & Large Displacement (E) & Small Displacement (E) & Large Angle Displacement (E) & Realistic Agent Size (T) & Realistic Goals (T) & Kicking Time (T) & No Ball Noise (N) & With Observation Noise (N) \\ \hline
        BS 1 & $\mathbf{0.88 \pm 0.1}$ & $0.87 \pm 0.1$ & $0.66 \pm 0.1$ & $0.62 \pm 0.1$ & $0.75 \pm 0.1$ & $0.71 \pm 0.1$ & $0.83 \pm 0.1$ & $0.54 \pm 0.1$ & $0.39 \pm 0.1$ & $0.74 \pm 0.1$ \\
        BS 2 & $\mathbf{1.00 \pm 0.0}$ & $\mathbf{1.00 \pm 0.0}$ & $0.89 \pm 0.1$ & $0.74 \pm 0.1$ & $0.99 \pm 0.0$ & $0.82 \pm 0.1$ & $0.98 \pm 0.0$ & $0.97 \pm 0.0$ & $0.96 \pm 0.0$ & $0.88 \pm 0.1$ \\
        BS 3 & $0.97 \pm 0.0$ & $\mathbf{0.99 \pm 0.0}$ & $0.80 \pm 0.1$ & $0.51 \pm 0.1$ & $0.71 \pm 0.1$ & $0.76 \pm 0.1$ & $0.92 \pm 0.1$ & $0.11 \pm 0.1$ & $0.50 \pm 0.1$ & $0.80 \pm 0.1$ \\
        D 1 & $\mathbf{0.84 \pm 0.1}$ & $0.69 \pm 0.1$ & $0.44 \pm 0.1$ & $0.00 \pm 0.0$ & $0.01 \pm 0.0$ & $0.05 \pm 0.0$ & $0.13 \pm 0.1$ & $0.33 \pm 0.1$ & $0.02 \pm 0.0$ & $0.12 \pm 0.1$ \\
        D 2 & $0.79 \pm 0.1$ & $\mathbf{0.93 \pm 0.1}$ & $0.21 \pm 0.1$ & $0.04 \pm 0.0$ & $0.81 \pm 0.1$ & $0.68 \pm 0.1$ & $0.20 \pm 0.1$ & $0.82 \pm 0.1$ & $0.89 \pm 0.1$ & $0.13 \pm 0.1$ \\
        D 3 & $0.80 \pm 0.1$ & $\mathbf{0.98 \pm 0.0}$ & $0.66 \pm 0.1$ & $0.00 \pm 0.0$ & $0.71 \pm 0.1$ & $0.46 \pm 0.1$ & $0.53 \pm 0.1$ & $0.65 \pm 0.1$ & $0.18 \pm 0.1$ & $0.46 \pm 0.1$ \\
        \hline
        \end{tabularx}
        \caption{Success Rates for Basic and Defender Simulation Experiments. Higher is better.}
    \end{subfigure}

    \vspace{3mm}

    \begin{subfigure}[b]{\linewidth}
        \centering
        \scriptsize
        \arrayrulecolor{gray!80} 
        \rowcolors{2}{white}{gray!25} 

        \begin{tabularx}{\linewidth}{|p{0.5cm}|XXXXXXXXXX|}
        \hline
        & Full MARL (F) & BHuman (F) & Large Displacement (E) & Small Displacement (E) & Large Angle Displacement (E) & Realistic Agent Size (T) & Realistic Goals (T) & Kicking Time (T) & No Ball Noise (N) & With Observation Noise (N) \\ \hline
        BS 1 & $36.50 \pm 1.7$ & $\mathbf{31.42 \pm 1.7}$ & $42.04 \pm 1.5$ & $41.62 \pm 2.0$ & $42.66 \pm 1.8$ & $32.65 \pm 1.6$ & $36.55 \pm 1.7$ & $45.48 \pm 2.3$ & $49.15 \pm 2.1$ & $39.45 \pm 1.9$ \\
        BS 2 & $16.42 \pm 1.3$ & $\mathbf{11.28 \pm 0.7}$ & $18.30 \pm 1.3$ & $20.96 \pm 1.5$ & $17.48 \pm 0.6$ & $21.45 \pm 1.6$ & $19.78 \pm 1.1$ & $23.65 \pm 1.8$ & $24.07 \pm 1.4$ & $19.47 \pm 1.5$ \\
        BS 3 & $33.56 \pm 1.2$ & $\mathbf{21.34 \pm 1.1}$ & $38.94 \pm 2.3$ & $37.22 \pm 2.8$ & $42.65 \pm 2.2$ & $41.05 \pm 1.7$ & $46.62 \pm 1.1$ & $52.45 \pm 4.3$ & $51.96 \pm 1.6$ & $34.00 \pm 1.5$ \\
        D 1 & $34.14 \pm 1.7$ & $\mathbf{29.79 \pm 2.1}$ & $45.06 \pm 2.9$ & $37.22 \pm 2.8$ & $57.00 \pm 0.0$ & $52.80 \pm 5.4$ & $43.69 \pm 4.1$ & $45.58 \pm 3.4$ & $54.00 \pm 2.0$ & $43.58 \pm 3.9$ \\
        D 2 & $34.74 \pm 2.5$ & $\mathbf{14.23 \pm 1.4}$ & $32.81 \pm 4.0$ & $59.00 \pm 1.1$ & $30.99 \pm 2.0$ & $52.01 \pm 1.6$ & $36.95 \pm 4.4$ & $43.22 \pm 2.0$ & $46.27 \pm 1.9$ & $36.54 \pm 6.0$ \\
        D 3 & $33.23 \pm 1.9$ & $\mathbf{20.40 \pm 1.0}$ & $40.66 \pm 3.1$ & $37.22 \pm 2.8$ & $34.22 \pm 2.1$ & $47.93 \pm 2.7$ & $29.85 \pm 2.0$ & $47.22 \pm 2.1$ & $49.56 \pm 3.8$ & $28.39 \pm 2.0$ \\
        \hline
        \end{tabularx}
        \caption{Time to Score for Basic and Defender Simulation Experiments. Lower is better.}
    \end{subfigure}

    \caption{Simulation experiment results. F are full methods. E are fidelity enhancements. T are training optimizations. N are noise. Our full method is Full MARL. Each column provides an experiment location for either the Basic Soccer (BS) task or the Defender (D) task. Number of trials is 100. 95\% confidence intervals computed using the Student t-distribution. }
    \label{fig:simulation_robot_results}

    \vspace{-1mm}
\end{figure*}

\section{Experimental Analysis}
\label{sec:setup}

We now present an empirical study designed to answer the questions:
\begin{enumerate}
    \item Can multi-agent robot control policies trained in an abstract simulator transfer directly to physical collaboration tasks?
    \item Which parts of our methodology are most critical for enabling transfer?
\end{enumerate}

\subsection{Empirical Setup}

To answer our empirical questions, we train policies in abstract simulation and then evaluate them on both physical robots and using a high-fidelity simulation as a surrogate for real world evaluation. 
In this subsection we detail the training setup and provide empirical results.

We consider our fullMARL training method to be: tuned agent displacement, large training agents, small training goals, no kicking time and ball-robot contact noise. To understand the impact of each component of our fullMARL method, we conduct an ablation study where we remove one component while keeping the other components unchanged.

\begin{figure}[t]
\vspace{3mm}
    \begin{subfigure}[t]{0.49\linewidth}
        \centering
        \includegraphics[width=\linewidth]{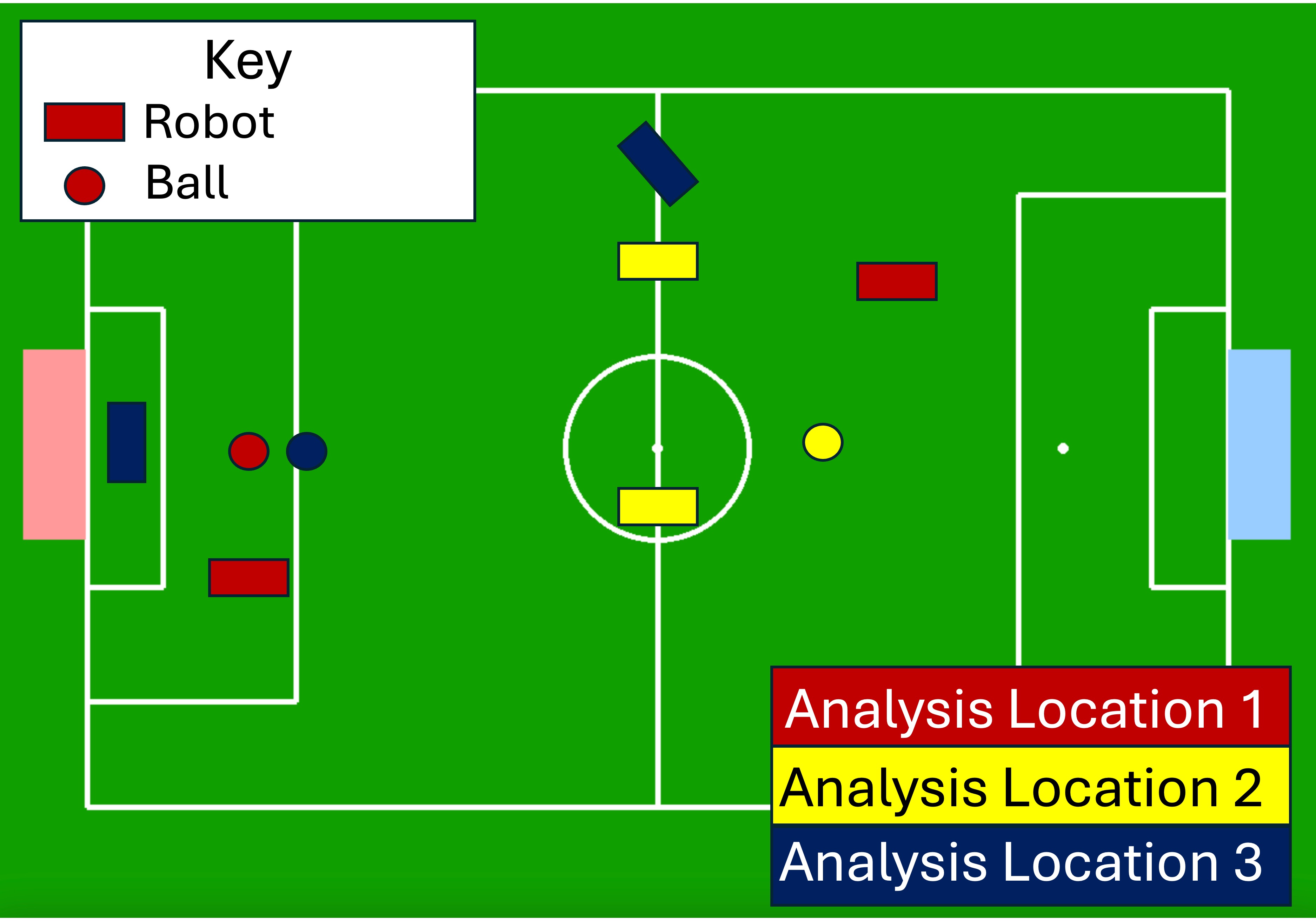}
    \end{subfigure}
    \begin{subfigure}[t]{0.49\linewidth}
        \centering
        \includegraphics[width=\linewidth]{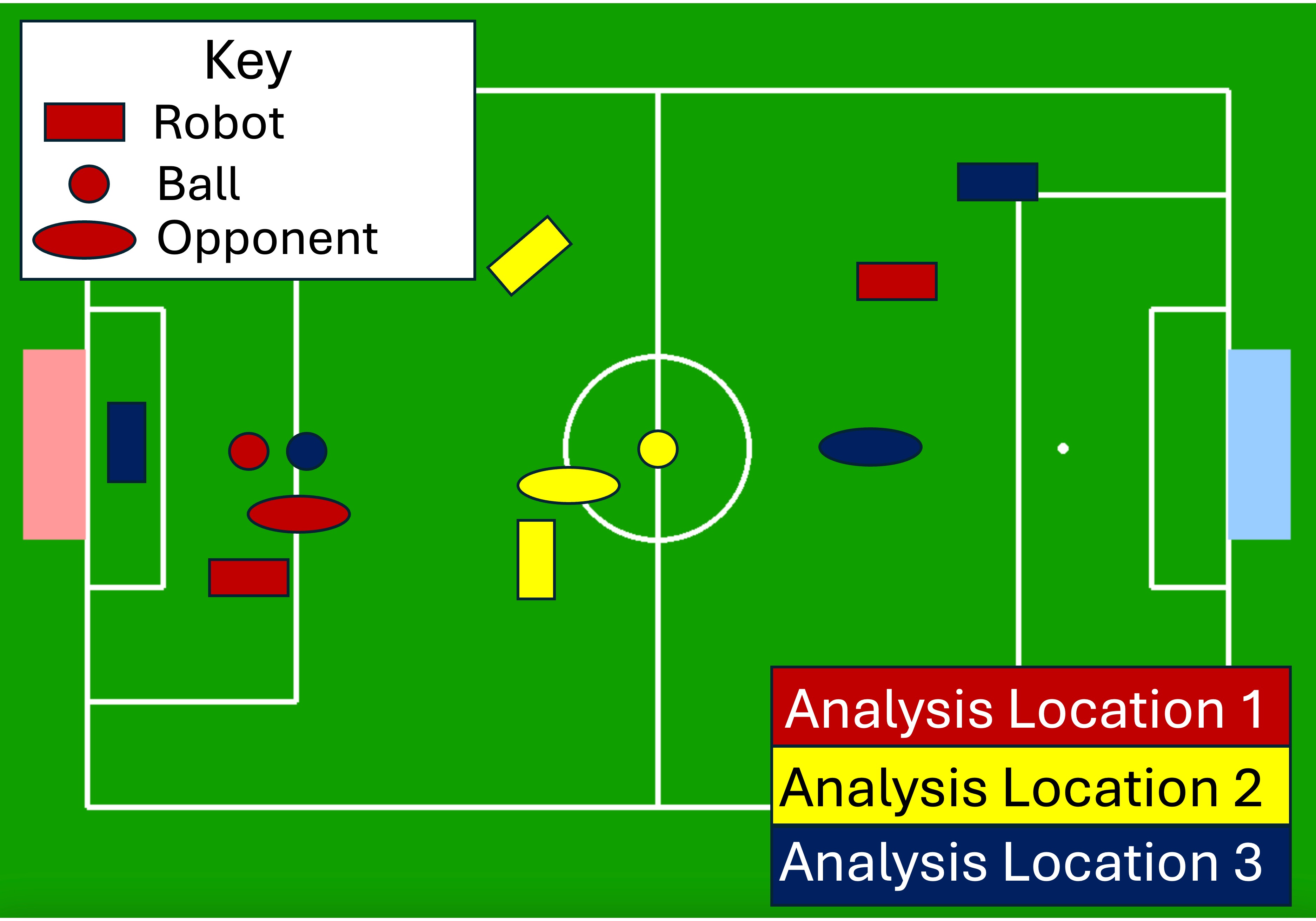}
    \end{subfigure}
    \caption{Analysis locations for basic soccer task and defender task. (Left) Basic soccer task. (Right) Static defender task. }
    \label{fig:allAnalysisLocations}
    \vspace{-3mm}
\end{figure}

For testing, we construct two cooperative tasks based on the robot soccer domain and measure performance based on success rate and time to score:
\begin{enumerate}
    \item \textbf{Basic Soccer.} Our first task, which we call basic soccer, has two robots and a ball on an empty field. During evaluation, we configure the start location of the robots such that performance is higher when they cooperate.
    \item \textbf{Static Defender.} The second task adds a non-moving defender robot that is set either between a robot and the ball or a robot and the goal. The addition of this robot introduces an obstacle that makes it more difficult to learn cooperative behaviors.
\end{enumerate}
In both tasks, we consider three initial configurations for evaluation which emphasize cooperative behaviors (\Cref{fig:allAnalysisLocations}).

On the physical robots we run the our suite of experiments with our fullMARL method along with one method from each ingredient category (\Cref{sec:methodology}). Physical robot experiments are costly to run, so we supplement them with high-fidelity simulation experiments. The high-fidelity simulation, developed by the B-Human team, closely matches the real-world performance of our policies (\Cref{fig:SimRobot}). In simulation we also compare our results to a RoboCup-winning architecture developed by B-Human, the same group that created the underlying localization and movement modules. We do note that while our policies perform at the level of the B-Human behavior in our testing, our evaluation is just one step toward robust behaviors that handle the complexities of robot soccer as effectively as their code does.

For policy training we use independent PPO \cite{stable-baselines3, schrittwieser2020mastering, SuperSuit} as the MARL approach in the abstract simulator and policies are trained with shared neural network weights which speeds up training time \cite{foerster2016learning}.

\subsection{Experimental Results}
\label{sec:experimentalResults}

Here we analyze our real robot and simulation experiment results.

\subsubsection{Simulation Fidelity Enhancement Analysis}

In both simulations and physical robots, we observe that an imprecise training walk speed negatively impacts the success rate. Specifically, the large training displacement caused the robot to score 0\% of the time in the static defender analysis setting and qualitatively the robot kicked the ball out of bounds on every run. In the static defender simulation, both small training displacement and large angle displacement performed 40\% worse than the tuned training displacement, achieving about a 0\% success rate. Overall, all changes to the displacement lowered the success rate by at least 15\% and at most 80\%. These findings show the importance of careful tuning of parameters to ensure the simulated motion closely matches the real-world motion dynamics.

\subsubsection{Training Optimization Analysis}

In our training optimization analysis, the large agent size performed significantly better than the correct size in the static defender task, likely because the large agents encouraged the robot to learn to maintain a farther distance to other robots. In simulation, the realistic agents, realistic goal sizes performed well on the basic soccer task but very poorly on the static defender task. This suggests that the changes to these parameters help more in the challenging scenarios. Lastly, for the no kicking time modification, we see less than a 50\% success rate in both the basic soccer and static defender tasks. Qualitatively, this is due to the difficulty in learning to kick the ball and instead pushing it due to the kicking delay. These findings show the need for careful consideration of training optimizations due to their potential large impact for on-robot performance.

\subsubsection{Simulation Stochasticity Enhancement Analysis}

We consider two types of noise to add stochasticity to our simulation. The first and more impactful of the two is ball contact noise. Quantitatively, without this change, the policies failed 9 out of 10 times on both physical robot tasks. Qualitatively, it only kicked the ball once throughout all the trials. When observing in AbstractSim, the agents rarely kick which suggests that without contact noise, the agents overfit to the optimized dynamics of AbstractSim and learn that pushing the ball is a consistent method of scoring goals. This result holds in both simulation domains, as the method does not achieve above a 40\% success rate. A lack of observation noise is less impactful but does hurt performance. This suggests that our underlying localization module is robust enough that adding noise during training hurts understanding of the ball location.

These results demonstrate that multi-agent robot control policies can be successfully trained in an abstract simulator and transferred to physical robot collaboration tasks but only when the environment is correctly designed. The ablation study shows the critical importance of accurate simulation for motion and the correct selection of training optimizations and simulation stochasticity.


\section{Discussion and Future Work}
\label{sec:limitationFutureWork}

Our empirical study demonstrates the feasibility of using highly abstract simulations for MARL training of cooperative control policies for physical robots.
In order to cross the reality gap from the abstract simulator, we identified three categories of simulator and training modifications that enable transfer from abstract simulation to reality.
Our physical robot experiments revealed that ball noise was the most impactful change. The use of ball-robot contact noise likely helped the agent avoid overfitting to the simplified abstract dynamics and underscores the critical importance of introducing simulation stochasticity to coarsely model contact dynamics.

While all three categories contributed substantially to successful sim-to-real transfer, of particular interest are the Training Optimizations, which yielded some of the most intriguing results. These optimizations, counterintuitively, involved reducing certain aspects of simulation realism to enhance real-world performance. This approach not only improved transfer success but also challenged our assumptions about the relationship between simulation fidelity and real-world effectiveness.

While a promising initial step toward the use of abstract simulators to develop multi-robot systems, we identify several limitations and directions for future work. First, the scenarios we used involved two robots cooperating and only required basic levels of teamwork. In the future, we plan to extend the use of abstract simulation for MARL training of teams of 5 robots to play against 5 other robots as used in the RoboCup SPL competition. This direction will require the study of sim2real in the presence of adversarial agents. Second, our method relies on existing lower level perception, localization, and control modules. The success of the decision making is contingent on the accuracy of the localization and movement control of the agent. This limitation is seen in the failure cases of our Defender task. The robot is unable to detect the defending robot and this results in a failed run. This limitation is increasingly less of an issue as many robotics systems ship with competent movement and perception code. Third, in this paper we manually selected the fidelity and training improvements for our specific task. Future work may aim to systematically identify the changes required for general abstract simulators to maximize on-robot performance. Finally, future work should consider the theoretical foundation of what creates a training or fidelity improvement. This foundation would provide deeper insight into the understanding of why some simulation aspects improve performance with less accurate world-modeling.


\section{Conclusion}
\label{sec:conclusion}

In this work we presented a method to enable teamwork on physical robots using MARL and abstract simulation. Starting with a base abstract simulator, we identified key simulator adjustments that enabled effective MARL and subsequent sim2real transfer. We conducted an extensive ablation study to identify the key components that significantly impact the success of our teamwork in the domain of robot soccer. These components fall into three categories: fidelity enhancement, training optimizations, and simulation stochasticity.  Our findings demonstrate that our method successfully enables teamwork on physical robots in the robot soccer domains and compares favorably to a state-of-the-art behavior architecture designed for the RoboCup competition. These findings show that MARL can be an effective tool for developing cooperative robot behaviors even without the use of high-fidelity simulation.


\section*{ACKNOWLEDGMENT}
Josiah Hanna acknowledges support from NSF (IIS-2410981), American Family Insurance through a research partnership with the University of Wisconsin—Madison’s Data Science Institute, the Wisconsin Alumni Research Foundation, and Sandia National Labs through a University Partnership Award.

\clearpage
\bibliographystyle{IEEEtran}
\bibliography{references}  

\begin{thebibliography}{10}
\providecommand{\url}[1]{#1}
\csname url@samestyle\endcsname
\providecommand{\newblock}{\relax}
\providecommand{\bibinfo}[2]{#2}
\providecommand{\BIBentrySTDinterwordspacing}{\spaceskip=0pt\relax}
\providecommand{\BIBentryALTinterwordstretchfactor}{4}
\providecommand{\BIBentryALTinterwordspacing}{\spaceskip=\fontdimen2\font plus
\BIBentryALTinterwordstretchfactor\fontdimen3\font minus \fontdimen4\font\relax}
\providecommand{\BIBforeignlanguage}[2]{{%
\expandafter\ifx\csname l@#1\endcsname\relax
\typeout{** WARNING: IEEEtran.bst: No hyphenation pattern has been}%
\typeout{** loaded for the language `#1'. Using the pattern for}%
\typeout{** the default language instead.}%
\else
\language=\csname l@#1\endcsname
\fi
#2}}
\providecommand{\BIBdecl}{\relax}
\BIBdecl

\bibitem{pitman2012cooperative}
R.~L. Pitman and J.~W. Durban, ``Cooperative hunting behavior, prey selectivity and prey handling by pack ice killer whales (orcinus orca), type b, in antarctic peninsula waters,'' \emph{Marine Mammal Science}, vol.~28, no.~1, pp. 16--36, 2012.

\bibitem{momeni2023multi}
M.~Momeni, H.~Soleimani, S.~Shahparvari, and B.~Afshar-Nadjafi, ``A multi-agency coordination resource allocation and routing decision-making problem: A coordinated truck-and-drone dss for improved wildfire detection coverage,'' \emph{International journal of disaster risk reduction}, vol.~97, p. 104027, 2023.

\bibitem{silver2016mastering}
D.~Silver, A.~Huang, C.~J. Maddison, A.~Guez, L.~Sifre, G.~Van Den~Driessche, J.~Schrittwieser, I.~Antonoglou, V.~Panneershelvam, M.~Lanctot \emph{et~al.}, ``Mastering the game of go with deep neural networks and tree search,'' \emph{nature}, vol. 529, no. 7587, pp. 484--489, 2016.

\bibitem{silver2017mastering}
D.~Silver, J.~Schrittwieser, K.~Simonyan, I.~Antonoglou, A.~Huang, A.~Guez, T.~Hubert, L.~Baker, M.~Lai, A.~Bolton \emph{et~al.}, ``Mastering the game of go without human knowledge,'' \emph{nature}, vol. 550, no. 7676, pp. 354--359, 2017.

\bibitem{schrittwieser2020mastering}
J.~Schrittwieser, I.~Antonoglou, T.~Hubert, K.~Simonyan, L.~Sifre, S.~Schmitt, A.~Guez, E.~Lockhart, D.~Hassabis, T.~Graepel \emph{et~al.}, ``Mastering atari, go, chess and shogi by planning with a learned model,'' \emph{Nature}, vol. 588, no. 7839, pp. 604--609, 2020.

\bibitem{vinyals2019grandmaster}
O.~Vinyals, I.~Babuschkin, W.~M. Czarnecki, M.~Mathieu, A.~Dudzik, J.~Chung, D.~H. Choi, R.~Powell, T.~Ewalds, P.~Georgiev \emph{et~al.}, ``Grandmaster level in starcraft ii using multi-agent reinforcement learning,'' \emph{Nature}, vol. 575, no. 7782, pp. 350--354, 2019.

\bibitem{berner2019dota}
C.~Berner, G.~Brockman, B.~Chan, V.~Cheung, P.~D{k{e}}biak, C.~Dennison, D.~Farhi, Q.~Fischer, S.~Hashme, C.~Hesse \emph{et~al.}, ``Dota 2 with large scale deep reinforcement learning,'' \emph{arXiv preprint arXiv:1912.06680}, 2019.

\bibitem{baker2019emergent}
B.~Baker, I.~Kanitscheider, T.~Markov, Y.~Wu, G.~Powell, B.~McGrew, and I.~Mordatch, ``Emergent tool use from multi-agent autocurricula,'' \emph{arXiv preprint arXiv:1909.07528}, 2019.

\bibitem{makoviychuk2021isaac}
V.~Makoviychuk, L.~Wawrzyniak, Y.~Guo, M.~Lu, K.~Storey, M.~Macklin, D.~Hoeller, N.~Rudin, A.~Allshire, A.~Handa \emph{et~al.}, ``Isaac gym: High performance gpu-based physics simulation for robot learning,'' \emph{arXiv preprint arXiv:2108.10470}, 2021.

\bibitem{todorov2012mujoco}
E.~Todorov, T.~Erez, and Y.~Tassa, ``Mujoco: A physics engine for model-based control,'' in \emph{2012 IEEE/RSJ international conference on intelligent robots and systems}.\hskip 1em plus 0.5em minus 0.4em\relax IEEE, 2012, pp. 5026--5033.

\bibitem{serban2018chrono}
R.~Serban, A.~Tasora, and D.~Negrut, ``Chrono: An open-source multi-physics simulation package,'' in \emph{presentado en The 5th Joint International Conference on Multibody System Dynamics, Lisboa, Portugal}, 2018.

\bibitem{ho2022people}
M.~K. Ho, D.~Abel, C.~G. Correa, M.~L. Littman, J.~D. Cohen, and T.~L. Griffiths, ``People construct simplified mental representations to plan,'' \emph{Nature}, vol. 606, no. 7912, pp. 129--136, 2022.

\bibitem{li2006towards}
L.~Li, T.~J. Walsh, and M.~L. Littman, ``Towards a unified theory of state abstraction for mdps.'' \emph{AI\&M}, vol.~1, no.~2, p.~3, 2006.

\bibitem{provost2004self}
J.~Provost, B.~J. Kuipers, and R.~Miikkulainen, ``Self-organizing perceptual and temporal abstraction for robot reinforcement learning,'' in \emph{AAAI Workshop on Learning and Planning in Markov Processes}, 2004, pp. 79--84.

\bibitem{shibata2006learning}
K.~Shibata, ``Learning of deterministic exploration and temporal abstraction in reinforcement learning,'' in \emph{2006 SICE-ICASE International Joint Conference}.\hskip 1em plus 0.5em minus 0.4em\relax IEEE, 2006, pp. 4569--4574.

\bibitem{ito2007autonomous}
K.~Ito, Y.~Fukumori, and A.~Takayama, ``Autonomous control of real snake-like robot using reinforcement learning; abstraction of state-action space using properties of real world,'' in \emph{2007 3rd International Conference on Intelligent Sensors, Sensor Networks and Information}.\hskip 1em plus 0.5em minus 0.4em\relax IEEE, 2007, pp. 389--394.

\bibitem{frommberger2010structural}
L.~Frommberger and D.~Wolter, ``Structural knowledge transfer by spatial abstraction for reinforcement learning agents,'' \emph{Adaptive Behavior}, vol.~18, no.~6, pp. 507--525, 2010.

\bibitem{ito2016reinforcement}
K.~Ito and Y.~Takeuchi, ``Reinforcement learning in dynamic environment: abstraction of state-action space utilizing properties of the robot body and environment,'' \emph{Artificial Life and Robotics}, vol.~21, pp. 11--17, 2016.

\bibitem{veerapaneni2020entity}
R.~Veerapaneni, J.~D. Co-Reyes, M.~Chang, M.~Janner, C.~Finn, J.~Wu, J.~Tenenbaum, and S.~Levine, ``Entity abstraction in visual model-based reinforcement learning,'' in \emph{Conference on Robot Learning}.\hskip 1em plus 0.5em minus 0.4em\relax PMLR, 2020, pp. 1439--1456.

\bibitem{agarwal2023legged}
A.~Agarwal, A.~Kumar, J.~Malik, and D.~Pathak, ``Legged locomotion in challenging terrains using egocentric vision,'' in \emph{Conference on robot learning}.\hskip 1em plus 0.5em minus 0.4em\relax PMLR, 2023, pp. 403--415.

\bibitem{geles2024demonstrating}
I.~Geles, L.~Bauersfeld, A.~Romero, J.~Xing, and D.~Scaramuzza, ``Demonstrating agile flight from pixels without state estimation,'' \emph{arXiv preprint arXiv:2406.12505}, 2024.

\bibitem{truong2023rethinking}
J.~Truong, M.~Rudolph, N.~H. Yokoyama, S.~Chernova, D.~Batra, and A.~Rai, ``Rethinking sim2real: Lower fidelity simulation leads to higher sim2real transfer in navigation,'' in \emph{Conference on Robot Learning}.\hskip 1em plus 0.5em minus 0.4em\relax PMLR, 2023, pp. 859--870.

\bibitem{zhang2024back}
Y.~Zhang, Y.~Hu, Y.~Song, D.~Zou, and W.~Lin, ``Back to newton's laws: Learning vision-based agile flight via differentiable physics,'' \emph{arXiv preprint arXiv:2407.10648}, 2024.

\bibitem{stulp2011hierarchical}
F.~Stulp and S.~Schaal, ``Hierarchical reinforcement learning with movement primitives,'' in \emph{2011 11th IEEE-RAS International Conference on Humanoid Robots}.\hskip 1em plus 0.5em minus 0.4em\relax IEEE, 2011, pp. 231--238.

\bibitem{ma2020hierarchical}
A.~Ma, M.~Ouimet, and J.~Cort{\'e}s, ``Hierarchical reinforcement learning via dynamic subspace search for multi-agent planning,'' \emph{Autonomous Robots}, vol.~44, no.~3, pp. 485--503, 2020.

\bibitem{nachum2019multi}
O.~Nachum, M.~Ahn, H.~Ponte, S.~Gu, and V.~Kumar, ``Multi-agent manipulation via locomotion using hierarchical sim2real,'' \emph{arXiv preprint arXiv:1908.05224}, 2019.

\bibitem{cutler2015real}
M.~Cutler, T.~J. Walsh, and J.~P. How, ``Real-world reinforcement learning via multifidelity simulators,'' \emph{IEEE Transactions on Robotics}, vol.~31, no.~3, pp. 655--671, 2015.

\bibitem{qiu2021low}
J.~Qiu, C.~Yu, W.~Liu, T.~Yang, J.~Yu, Y.~Wang, and H.~Yang, ``Low-cost multi-agent navigation via reinforcement learning with multi-fidelity simulator,'' \emph{IEEE Access}, vol.~9, pp. 84\,773--84\,782, 2021.

\bibitem{beard2022black}
J.~J. Beard and A.~Baheri, ``Black-box safety validation of autonomous systems: A multi-fidelity reinforcement learning approach,'' \emph{arXiv preprint arXiv:2203.03451}, 2022.

\bibitem{agrawal2023reinforcement}
A.~Agrawal and C.~McComb, ``Reinforcement learning for efficient design space exploration with variable fidelity analysis models,'' \emph{Journal of Computing and Information Science in Engineering}, vol.~23, no.~4, p. 041004, 2023.

\bibitem{bhola2023multi}
S.~Bhola, S.~Pawar, P.~Balaprakash, and R.~Maulik, ``Multi-fidelity reinforcement learning framework for shape optimization,'' \emph{Journal of Computational Physics}, vol. 482, p. 112018, 2023.

\bibitem{leguizamo2022deep}
D.~F. Leguizamo, H.-J. Yang, X.~Y. Lee, and S.~Sarkar, ``Deep reinforcement learning for robotic control with multi-fidelity models,'' \emph{IFAC-PapersOnLine}, vol.~55, no.~37, pp. 193--198, 2022.

\bibitem{truong2021learning}
J.~Truong, D.~Yarats, T.~Li, F.~Meier, S.~Chernova, D.~Batra, and A.~Rai, ``Learning navigation skills for legged robots with learned robot embeddings,'' in \emph{2021 IEEE/RSJ International Conference on Intelligent Robots and Systems (IROS)}.\hskip 1em plus 0.5em minus 0.4em\relax IEEE, 2021, pp. 484--491.

\bibitem{zhao2020sim}
W.~Zhao, J.~P. Queralta, and T.~Westerlund, ``Sim-to-real transfer in deep reinforcement learning for robotics: a survey,'' in \emph{2020 IEEE symposium series on computational intelligence (SSCI)}.\hskip 1em plus 0.5em minus 0.4em\relax IEEE, 2020, pp. 737--744.

\bibitem{salvato2021crossing}
E.~Salvato, G.~Fenu, E.~Medvet, and F.~A. Pellegrino, ``Crossing the reality gap: A survey on sim-to-real transferability of robot controllers in reinforcement learning,'' \emph{IEEE Access}, vol.~9, pp. 153\,171--153\,187, 2021.

\bibitem{kaufmann2023champion}
E.~Kaufmann, L.~Bauersfeld, A.~Loquercio, M.~M{\"u}ller, V.~Koltun, and D.~Scaramuzza, ``Champion-level drone racing using deep reinforcement learning,'' \emph{Nature}, vol. 620, no. 7976, pp. 982--987, 2023.

\bibitem{hwangbo2017control}
J.~Hwangbo, I.~Sa, R.~Siegwart, and M.~Hutter, ``Control of a quadrotor with reinforcement learning,'' \emph{IEEE Robotics and Automation Letters}, vol.~2, no.~4, pp. 2096--2103, 2017.

\bibitem{cao2019target}
X.~Cao, C.~Sun, and M.~Yan, ``Target search control of auv in underwater environment with deep reinforcement learning,'' \emph{IEEE Access}, vol.~7, pp. 96\,549--96\,559, 2019.

\bibitem{carlucho2018adaptive}
I.~Carlucho, M.~De~Paula, S.~Wang, Y.~Petillot, and G.~G. Acosta, ``Adaptive low-level control of autonomous underwater vehicles using deep reinforcement learning,'' \emph{Robotics and Autonomous Systems}, vol. 107, pp. 71--86, 2018.

\bibitem{hester2013texplore}
T.~Hester and P.~Stone, ``Texplore: real-time sample-efficient reinforcement learning for robots,'' \emph{Machine learning}, vol.~90, pp. 385--429, 2013.

\bibitem{mahmood2018benchmarking}
A.~R. Mahmood, D.~Korenkevych, G.~Vasan, W.~Ma, and J.~Bergstra, ``Benchmarking reinforcement learning algorithms on real-world robots,'' in \emph{Conference on robot learning}.\hskip 1em plus 0.5em minus 0.4em\relax PMLR, 2018, pp. 561--591.

\bibitem{falco2018policy}
P.~Falco, A.~Attawia, M.~Saveriano, and D.~Lee, ``On policy learning robust to irreversible events: An application to robotic in-hand manipulation,'' \emph{IEEE Robotics and Automation Letters}, vol.~3, no.~3, pp. 1482--1489, 2018.

\bibitem{hazara2019transferring}
M.~Hazara and V.~Kyrki, ``Transferring generalizable motor primitives from simulation to real world,'' \emph{IEEE Robotics and Automation Letters}, vol.~4, no.~2, pp. 2172--2179, 2019.

\bibitem{chalvatzaki2019learn}
G.~Chalvatzaki, X.~S. Papageorgiou, P.~Maragos, and C.~S. Tzafestas, ``Learn to adapt to human walking: A model-based reinforcement learning approach for a robotic assistant rollator,'' \emph{IEEE Robotics and Automation Letters}, vol.~4, no.~4, pp. 3774--3781, 2019.

\bibitem{loquercio2021learning}
A.~Loquercio, E.~Kaufmann, R.~Ranftl, M.~M{\"u}ller, V.~Koltun, and D.~Scaramuzza, ``Learning high-speed flight in the wild,'' \emph{Science Robotics}, vol.~6, no.~59, p. eabg5810, 2021.

\bibitem{haarnoja2024learning}
T.~Haarnoja, B.~Moran, G.~Lever, S.~H. Huang, D.~Tirumala, J.~Humplik, M.~Wulfmeier, S.~Tunyasuvunakool, N.~Y. Siegel, R.~Hafner \emph{et~al.}, ``Learning agile soccer skills for a bipedal robot with deep reinforcement learning,'' \emph{Science Robotics}, vol.~9, no.~89, p. eadi8022, 2024.

\bibitem{tirumala2024learning}
D.~Tirumala, M.~Wulfmeier, B.~Moran, S.~Huang, J.~Humplik, G.~Lever, T.~Haarnoja, L.~Hasenclever, A.~Byravan, N.~Batchelor \emph{et~al.}, ``Learning robot soccer from egocentric vision with deep reinforcement learning,'' \emph{arXiv preprint arXiv:2405.02425}, 2024.

\bibitem{duan2012multi}
Y.~Duan, B.~X. Cui, and X.~H. Xu, ``A multi-agent reinforcement learning approach to robot soccer,'' \emph{Artificial Intelligence Review}, vol.~38, pp. 193--211, 2012.

\bibitem{park2001modular}
K.-H. Park, Y.-J. Kim, and J.-H. Kim, ``Modular q-learning based multi-agent cooperation for robot soccer,'' \emph{Robotics and Autonomous systems}, vol.~35, no.~2, pp. 109--122, 2001.

\bibitem{tobin2017domain}
J.~Tobin, R.~Fong, A.~Ray, J.~Schneider, W.~Zaremba, and P.~Abbeel, ``Domain randomization for transferring deep neural networks from simulation to the real world,'' in \emph{2017 IEEE/RSJ international conference on intelligent robots and systems (IROS)}.\hskip 1em plus 0.5em minus 0.4em\relax IEEE, 2017, pp. 23--30.

\bibitem{kaspar2020sim2real}
M.~Kaspar, J.~D.~M. Osorio, and J.~Bock, ``Sim2real transfer for reinforcement learning without dynamics randomization,'' in \emph{2020 IEEE/RSJ International Conference on Intelligent Robots and Systems (IROS)}.\hskip 1em plus 0.5em minus 0.4em\relax IEEE, 2020, pp. 4383--4388.

\bibitem{hanna2021grounded}
J.~P. Hanna, S.~Desai, H.~Karnan, G.~Warnell, and P.~Stone, ``Grounded action transformation for sim-to-real reinforcement learning,'' \emph{Machine Learning}, vol. 110, no.~9, pp. 2469--2499, 2021.

\bibitem{tan2018sim}
J.~Tan, T.~Zhang, E.~Coumans, A.~Iscen, Y.~Bai, D.~Hafner, S.~Bohez, and V.~Vanhoucke, ``Sim-to-real: Learning agile locomotion for quadruped robots,'' \emph{arXiv preprint arXiv:1804.10332}, 2018.

\bibitem{blumenkamp2022framework}
J.~Blumenkamp, S.~Morad, J.~Gielis, Q.~Li, and A.~Prorok, ``A framework for real-world multi-robot systems running decentralized gnn-based policies,'' in \emph{2022 International Conference on Robotics and Automation (ICRA)}.\hskip 1em plus 0.5em minus 0.4em\relax IEEE, 2022, pp. 8772--8778.

\bibitem{de2020deep}
C.~S. de~Witt, B.~Peng, P.-A. Kamienny, P.~Torr, W.~B{\"o}hmer, and S.~Whiteson, ``Deep multi-agent reinforcement learning for decentralized continuous cooperative control,'' \emph{arXiv preprint arXiv:2003.06709}, vol.~19, 2020.

\bibitem{cui2024distributed}
Z.~Cui, C.~Yang, and X.~Cao, ``A distributed simulation-to-reality transfer framework for multi-agent reinforcement learning,'' in \emph{2024 36th Chinese Control and Decision Conference (CCDC)}.\hskip 1em plus 0.5em minus 0.4em\relax IEEE, 2024, pp. 3807--3812.

\bibitem{candela2022transferring}
E.~Candela, L.~Parada, L.~Marques, T.-A. Georgescu, Y.~Demiris, and P.~Angeloudis, ``Transferring multi-agent reinforcement learning policies for autonomous driving using sim-to-real,'' in \emph{2022 IEEE/RSJ International Conference on Intelligent Robots and Systems (IROS)}.\hskip 1em plus 0.5em minus 0.4em\relax IEEE, 2022, pp. 8814--8820.

\bibitem{zhao2020towards}
W.~Zhao, J.~P. Queralta, L.~Qingqing, and T.~Westerlund, ``Towards closing the sim-to-real gap in collaborative multi-robot deep reinforcement learning,'' in \emph{2020 5th International conference on robotics and automation engineering (ICRAE)}.\hskip 1em plus 0.5em minus 0.4em\relax IEEE, 2020, pp. 7--12.

\bibitem{djeumou2023learn}
F.~Djeumou, C.~Neary, and U.~Topcu, ``How to learn and generalize from three minutes of data: Physics-constrained and uncertainty-aware neural stochastic differential equations,'' \emph{arXiv preprint arXiv:2306.06335}, 2023.

\bibitem{huang2022trade}
J.~Huang, Y.~Zhang, F.~Giardina, and A.~Rosendo, ``Trade-off on sim2real learning: Real-world learning faster than simulations,'' in \emph{2022 8th International Conference on Control, Automation and Robotics (ICCAR)}.\hskip 1em plus 0.5em minus 0.4em\relax IEEE, 2022, pp. 95--100.

\bibitem{wu2023daydreamer}
P.~Wu, A.~Escontrela, D.~Hafner, P.~Abbeel, and K.~Goldberg, ``Daydreamer: World models for physical robot learning,'' in \emph{Conference on Robot Learning}.\hskip 1em plus 0.5em minus 0.4em\relax PMLR, 2023, pp. 2226--2240.

\bibitem{thuruthel2018model}
T.~G. Thuruthel, E.~Falotico, F.~Renda, and C.~Laschi, ``Model-based reinforcement learning for closed-loop dynamic control of soft robotic manipulators,'' \emph{IEEE Transactions on Robotics}, vol.~35, no.~1, pp. 124--134, 2018.

\bibitem{zhang2019solar}
M.~Zhang, S.~Vikram, L.~Smith, P.~Abbeel, M.~Johnson, and S.~Levine, ``Solar: Deep structured representations for model-based reinforcement learning,'' in \emph{International conference on machine learning}.\hskip 1em plus 0.5em minus 0.4em\relax PMLR, 2019, pp. 7444--7453.

\bibitem{hester2012rtmba}
T.~Hester, M.~Quinlan, and P.~Stone, ``Rtmba: A real-time model-based reinforcement learning architecture for robot control,'' in \emph{2012 IEEE International Conference on Robotics and Automation}.\hskip 1em plus 0.5em minus 0.4em\relax IEEE, 2012, pp. 85--90.

\bibitem{becker2020learning}
P.~Becker-Ehmck, M.~Karl, J.~Peters, and P.~van~der Smagt, ``Learning to fly via deep model-based reinforcement learning,'' \emph{arXiv preprint arXiv:2003.08876}, 2020.

\bibitem{martinez2015safe}
D.~Mart{\'\i}nez, G.~Alenya, and C.~Torras, ``Safe robot execution in model-based reinforcement learning,'' in \emph{2015 IEEE/RSJ International Conference on Intelligent Robots and Systems (IROS)}.\hskip 1em plus 0.5em minus 0.4em\relax IEEE, 2015, pp. 6422--6427.

\bibitem{lambert2019low}
N.~O. Lambert, D.~S. Drew, J.~Yaconelli, S.~Levine, R.~Calandra, and K.~S. Pister, ``Low-level control of a quadrotor with deep model-based reinforcement learning,'' \emph{IEEE Robotics and Automation Letters}, vol.~4, no.~4, pp. 4224--4230, 2019.

\bibitem{imanberdiyev2016autonomous}
N.~Imanberdiyev, C.~Fu, E.~Kayacan, and I.-M. Chen, ``Autonomous navigation of uav by using real-time model-based reinforcement learning,'' in \emph{2016 14th international conference on control, automation, robotics and vision (ICARCV)}.\hskip 1em plus 0.5em minus 0.4em\relax IEEE, 2016, pp. 1--6.

\bibitem{van2021model}
P.~Van Der~Vaart, A.~Mahajan, and S.~Whiteson, ``Model based multi-agent reinforcement learning with tensor decompositions,'' \emph{arXiv preprint arXiv:2110.14524}, 2021.

\bibitem{zhang2020model}
K.~Zhang, S.~Kakade, T.~Basar, and L.~Yang, ``Model-based multi-agent rl in zero-sum markov games with near-optimal sample complexity,'' \emph{Advances in Neural Information Processing Systems}, vol.~33, pp. 1166--1178, 2020.

\bibitem{pasztor2021efficient}
B.~Pasztor, I.~Bogunovic, and A.~Krause, ``Efficient model-based multi-agent mean-field reinforcement learning,'' \emph{arXiv preprint arXiv:2107.04050}, 2021.

\bibitem{sessa2022efficient}
P.~G. Sessa, M.~Kamgarpour, and A.~Krause, ``Efficient model-based multi-agent reinforcement learning via optimistic equilibrium computation,'' in \emph{International Conference on Machine Learning}.\hskip 1em plus 0.5em minus 0.4em\relax PMLR, 2022, pp. 19\,580--19\,597.

\bibitem{egorov2022scalable}
V.~Egorov and A.~Shpilman, ``Scalable multi-agent model-based reinforcement learning,'' \emph{arXiv preprint arXiv:2205.15023}, 2022.

\bibitem{bhuman_code_release}
\BIBentryALTinterwordspacing
B.-H. Team, ``{B-Human Code Release},'' \url{https://wiki.b-human.de/coderelease2023/}, October 2023. [Online]. Available: \url{https://wiki.b-human.de/coderelease2023/}
\BIBentrySTDinterwordspacing

\bibitem{labiosa2024reinforcement}
A.~Labiosa, Z.~Wang, S.~Agarwal, W.~Cong, G.~Hemkumar, A.~N. Harish, B.~Hong, J.~Kelle, C.~Li, Y.~Li \emph{et~al.}, ``Reinforcement learning within the classical robotics stack: A case study in robot soccer,'' \emph{arXiv preprint arXiv:2412.09417}, 2024.

\bibitem{smith2005open}
R.~Smith \emph{et~al.}, ``Open dynamics engine,'' 2005.

\bibitem{stable-baselines3}
\BIBentryALTinterwordspacing
A.~Raffin, A.~Hill, A.~Gleave, A.~Kanervisto, M.~Ernestus, and N.~Dormann, ``Stable-baselines3: Reliable reinforcement learning implementations,'' \emph{Journal of Machine Learning Research}, vol.~22, no. 268, pp. 1--8, 2021. [Online]. Available: \url{http://jmlr.org/papers/v22/20-1364.html}
\BIBentrySTDinterwordspacing

\bibitem{SuperSuit}
J.~K. Terry, B.~Black, and A.~Hari, ``Supersuit: Simple microwrappers for reinforcement learning environments,'' \emph{arXiv preprint arXiv:2008.08932}, 2020.

\bibitem{foerster2016learning}
J.~Foerster, I.~A. Assael, N.~De~Freitas, and S.~Whiteson, ``Learning to communicate with deep multi-agent reinforcement learning,'' \emph{Advances in neural information processing systems}, vol.~29, 2016.

\end{thebibliography}

\end{document}